\documentclass{llncs}

\usepackage{graphicx}
\begin{document}

\title{Personalized sentence generation using generative adversarial networks with author-specific word usage}

\author{
Chenhan Yuan and Yi-chin Huang*}
\institute{University of Electronic Science and Technology of China\\
           National Pingtung University*\\
           yuanchenhan@std.uestc.edu.cn, ychin.huang@gmail.com}

\maketitle

\begin{abstract}
The author-specific word usage is a vital feature to let readers perceive the writing style of the author. In this work, a personalized sentence generation method based on generative adversarial networks (GANs) is proposed to cope with this issue. The frequently used function word and content word are incorporated not only as the input features but also as the sentence structure constraint for the GAN training. For the sentence generation with the related topics decided by the user, the Named Entity Recognition (NER) information of the input words is also used in the network training. We compared the proposed method with the GAN-based sentence generation methods, and the experimental results showed that the generated sentences using our method are more similar to the original sentences of the same author based on the objective evaluation such as BLEU and SimHash score. 

\end{abstract}

\section{Introduction}
Text generation, as a basic task of natural language processing, has many applications, such as dialogue robots \cite{li2017adversarial}, machine translation \cite{hu2017controllable}, paraphrasing \cite{power2005automatic} and so on. In previous works, many researchers \cite{power2005automatic} extracted grammar rules from text and used them to generate new texts. These works are capable of generating semantically rich and grammatically correct text, but due to the fixed grammar rules, the differences between generated sentences are quite limited. With the rise of deep learning \cite{zhang2016generating,yu2017seqgan,guo2018long,lin2017adversarial}, researchers have tried to introduce neural networks to generate sentences. Long Short-Term Memory (LSTM \cite{hochreiter1997long}) is used as a sequential neural network model to generate sentences. It can judge the generation of the next word based on the words that have been generated before. The use of neural networks to generate text has greatly increased the variation of text. \\
\\
Lately, the Generative Adversarial Networks (GAN) \cite{goodfellow2014generative} has been introduced, and several variants of the GAN model for generating text have been proposed. These GAN variants yield good performances in the context of generating short texts, such as SeqGAN \cite{yu2017seqgan}, RankGAN \cite{lin2017adversarial}, TextGAN \cite{zhang2016generating}. To generate long text, in \cite{guo2018long}, they introduced LeakGAN to enable the discriminator leaks features extracted from its input to generator, which uses this signal to guide the outputs in each generation step. They built a hierarchical Reinforcement Learning (RL) \cite{sutton2000policy} framework as a generator, in which the module named MANAGER accepts the leaked feature to form a goal vector, which is help guiding another module named WORKER to generate the next word. \\
\\
However, one application of the text generation has not been explored extensively, which is text generation with personalized writing style. The personalized writing style could be helpful for various applications. For example, in the scenario of responding to an email or message automatically, if we want to let the receivers convinced that the reply is written by the person, the writing style of that email should be similar to that of the user. Another example would be generating a paragraph of a specific topic and is written by a well-known author. Therefore, we would like to let the user to define the topic of the generated text, which means the text generated by our method should be related to the user-defined topic and the collected texts form a specific author are used to extract the personalized writing style information as the author-specific feature, which is fed into the GAN framework to guide the GAN to generate text with personalized writing style and user defined topic.
\section{Related Work}
\subsection{Language Features}
The syntactic and content information from text is a long-standing topic in natural language processing. The syntactic structure of one sentence is often closely related to the part-of-speech (POS) sequence corresponding to that sentence; therefore, many efforts are made to improve the performance of labeling POS tag automatically. In \cite{toutanova2003feature}, they used cyclic dependency network to consider the POS context to obtain higher POS tagging accuracy. Named Entity Recognition (NER), which refers to the identification of entities with specific meaning in the text, is an important analysis process to obtain text content information. In order to improve the accuracy of NER process, another work \cite{finkel2005incorporating} is based on the conditional random field model, using Gibbs sampling to adopt the structure of long sentence instead of only local feature. In addition, the relationship between each word is also an important characteristic of text. Researchers attempt to represent words in text using vectors, so that the relationship between words can be determined by calculating the cosine similarity between vectors. The Word2vec algorithm is proposed in \cite{mikolov2013efficient}, which includes two architectures: Continuous Bag-of-Words Model and Continuous Skip-gram Model, to calculate continuous vector representations of words.

\subsection{Text Generation using GAN}
With the development of deep learning, many text generation works have begun to use neural networks in recent years. For example, Generative Adversarial Networks (GAN) \cite{goodfellow2014generative} provides a novel way to generate text, which consists of a generator and a discriminator. Discriminator determines whether the input data is real or generated, and Generator generates data and try to convince the discriminator that the generated data is real. However, in context of discrete data input, discriminator cannot propagate the gradient back to the generator as in standard GAN training. To address this problem to generate text via GAN, some researchers employ Long Short-term Memory (LSTM \cite{hochreiter1997long}) and convolutional neural network (CNN \cite{kim2014convolutional}) for adversarial training to generate realistic text and optimize a new feature distance when training the generator \cite{zhang2016generating}. A method is inspired by the Reinforcement Learning (RL)\cite{sutton2000policy} reward signal come from the GAN discriminator judged on a complete sequence, which is passed back to the intermediate state-action steps using Monte Carlo search \cite{yu2017seqgan}. To generate long text, LeakGAN \cite{guo2018long} is introduced to enable discriminator leak features extracted from its input to generator, which will use this signal to guide the outputs for each word generation step. They built a hierarchical RL framework as a generator, in which the module named MANAGER accepts the leaked feature to form a goal vector to guide another module named WORKER to generate the next word.\\
\\
Those variants of the GAN framework are helpful for text generation and the performance is superior than the conventional rule-based methods. Nevertheless, to collect a text corpus of a specific author with various topics is hard to achieve, since one author usually write only limited amount of articles with similar topics. Therefore, the main goal of the proposed method is try to use all texts collected from different authors for training the GAN model in order to let the covered topics as many as possible. More importantly, the generated sentences should be perceived as written by a specific author. We first implemented NER on all collected articles to determine which articles contain information related to user-defined topics and save them as the training set. Besides, the bigram language model of the POS tag and the corresponding structural words of the target author are extracted as the author’s personalized writing habit. These frequently used syntactic structure information is helpful for text generation module as the personalized features. Also, the frequently used words of each author are extracted to represent another part of personalized information. Here, the text generation module is inspired by the LeakGAN method, because we would like to let the generator use the personalized features to generate each word while the discriminator leaks the information of the previous generated words to the generator. In order to let the LeakGAN model to adopt the personalized feature for each epoch of word generation, we modified the structure of its CNN and the input and output of the generator.
\begin{figure}[h]
	\centering  
	\includegraphics[width=0.7\linewidth]{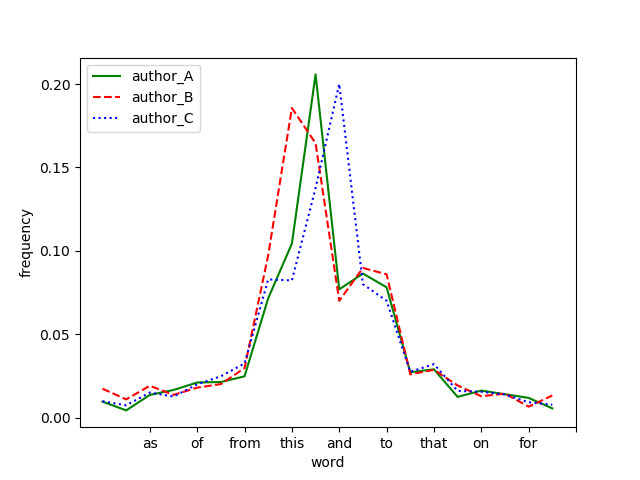}  
	\caption{The term frequency distributions of structure words from
different authors.}  
	\label{01}   
\end{figure}
\section{Proposed Method}
Generally speaking, the writing style of an author could be defined by the frequently used syntactic structure and the word usage. Here, we collected texts written by different authors and compare the term frequency of the structure words. As shown in Figure \ref{01}, the term frequency distributions of the structure words of different authors are quite different, which suggests that the structure word usage should be taken into account for generating sentences as written by a specific author.\\
\\
\begin{figure}[h]
	\centering  
	\includegraphics[width=0.7\linewidth]{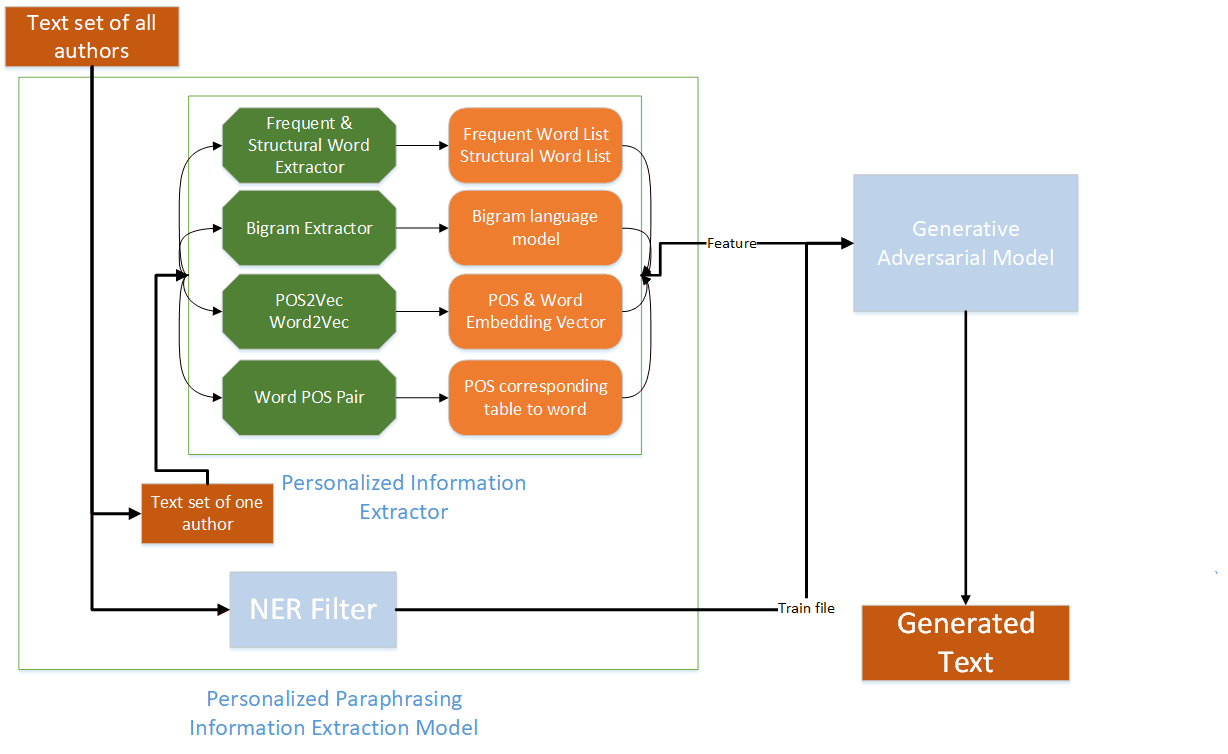}  
	\caption{The system framework of the proposed method for generating
personalized sentences.}  
	\label{02}   
\end{figure}
The system framework of the proposed method is depicted in Figure \ref{02}. First, the structural words and their corresponding part-of-speech (POS) information are extracted to construct a bigram grammar model to represent syntactic structures. Besides, the frequently used content words of the target author are extracted from the corresponding texts based on the term frequency. For guiding the GAN to generate the sentences, the list of structural words used by the author and his unique bigram grammar model will be fed into the discriminator. When the discriminator judging whether the sentence is real or generated, these information will be used.\\
\\
Then, the frequently used words of the target author and the syntactic structure features will be combined with the leaked feature of the discriminator as a new feature to feed the manager module of the generator to help guiding the next word generation. For the worker module of the generator, the word embedding vector generated by the previous step and its corresponding POS embedding vector will be combined as the input features. When generating the next word, the syntactic structure is considered to let the generated word sequence become more similar to the sentence structure that the target author prefer to use.\\
\\
Another vital feature of the proposed method is the user-defined topic. For the sentences of the generated text to be related to the user-defined topic, we implement named entity recognition (NER) to filter the training corpus. The texts of all author that are qualified for the NER filter will be used as a training corpus to train our model. This is because the NER filter can guarantee that the text used for training contains information related to the user-defined topic so that the generated text will also have relevant information. In the following section, the detail information of the personalized information and how to we applied the information to guild GAN is introduced.
\subsection{Personalized related Information Extraction
Model}
\subsubsection{NER FILTER}:\\
\\
In order to enable the generated sentences contain information related to the user-defined topic, NER filter is used to select all the texts that meet the requirements and are served as the training set. We perform named entity recognition (NER) on the text of all authors. When the named entity contained in the article is the same as the named entity decided by the user, the text is included in the training set. Because for the news corpus, the named entity contains most of the important information in an article, the training set ensures that the generated text will contain information related to the user’s topic. Compared to conventional generative adversarial networks, which directly adopt all collected text as a training corpus, although the number of texts decreases after passing the NER filter, the generated sentences are more related to a particular topic.
\subsubsection{PERSONALIZED INFORMATION EXTRACTOR}:\\
\\
In general, the personalized writing style mainly consists of specific word usage, which contains all the words the author prefer but may not commonly used by others, as well as the preferred syntactic structure of the author. For example, when writing an article, an author might prefer to use “film” instead of “movie.” An author may also like to use the “..., thus...” syntactic structure instead of “because..., ...”. Therefore, our approach extracts the personalized information from words and syntactic structures which is frequently used by an author. The detail personalized information extraction are described as follows.\\
\\
{\bfseries Frequent Word Extractor:} We calculate the ratio of the term frequency to determine whether a specific word is a frequently-used word for an author. Equation \ref{1} calculates the mean value of the i-th word in the collected texts. Equation \ref{2} then calculates the the ratio of the i-th word for each author. $TF_{word(i,j)}$ represents the term frequency of the i-th word written by j-th author in all texts. Then, a threshold is used to decided whether a word is a author preferred word or not. For example, if a threshold is set to 0.3, then when the calculated result of eq. \ref{2} is greater than 0.3, that means the i-th word is a preferred word for the j-th author. In such case, the greater result indicates that the j-th author prefers to use the i-th word than other authors. Otherwise, the i-th word is a common word, when means this word is used by each author.\\
\begin{equation}\label{1}
TF_{mean\_word\_i}=\frac{\sum_{j}\frac{TF_{word\_ij}}{\sum_{i}TF_{word\_ij}}}{j}
\end{equation}
\begin{equation}\label{2}
\frac{\frac{TF_{word\_ij}}{\sum_{i}TF_{word\_ij}}-TF_{mean\_word\_i}}{TF_{mean\_word\_i}}
\end{equation}
\\
{\bfseries Bigram Extractor:} In statistical machine translation, bigram language model \cite{zhai2004study} is a useful probability distribution, which preserves the phrase characteristics of the text. Since the training text set is decided by the NER filter, the author-specific syntactic structure is not preserved in such training set. Therefore, we adopt the bigram language model based on the structure word and its POS tag. The author-specific syntactic structure information could be defined by the POS tag rather than directly using the word information. The POS bigram probability distribution and the language model are extracted using Eq. \ref{3} where $POS_k$ represents the k-th bigram POS combination that exists in the texts. In order to deal with out-of-data problem in the bigram POS combination, we adopted the add-delta method when calculating the probability distribution. Here, the $EmptyPOS_l$ indicates the l-th bigram combination that does not occurred in the collected texts, and is the smoothing parameter, which is set to 0.5 based on our experimental result.\\
\begin{equation}\label{3}
POS_{bigram\_k}=\frac{POS_{k}}{\sum_{k}POS_{k}+\sum_{l}\delta\times{EmptyPOS_l}}
\end{equation}
\\
{\bfseries Word2Vec and POS2Vec:} In order to generate sentences containing personalized word usage of the target author, the cosine similarity of the word vectors is applied. Here we measure the distance between frequently used word of the target author and those words which are more commonly used among all authors in the collected training set. By applying this information to the generator of GAN framework, the proposed method replaces the suitable word based on algorithm described in Section 3.1. Instead of using the word embedding directly from the GAN model, we trained a word embedding by Word2Vec \cite{mikolov2013efficient} algorithm using the features extracted from the personalized related information extraction phase to determine word similarity. This would ensure the personalized word usage to guild the GAN training and could reducing the number of training epoch. Besides, the syntactic structure is represented using the POS vector and served as the part of the input for the generator. This is achieved by automatically labeling the POS tags of an author and then apply the word2vec algorithm to obtain the POS embedding.\\
\\
{\bfseries Word POS mapping:} Because each word vector in tensorflow \cite{abadi2016tensorflow} flows as a tensor in the graph, it is difficult to dynamically label POS on newly generated words while generating sentences. Therefore, we map each word to a unique POS at this stage. Based on the statistics of the training corpus, more than 80\% of the word can be labeled as a specific POS tag. Therefore, our operation will not only not adversely affect the results, but also avoid passing the error generated in labeling POS on word process to GAN. Finally, we got a POS mapping table of each word to feed into GAN.
\subsection{Modified Generative Adversarial Networks Model}
The proposed GAN-based method is inspired by LeakGAN with several modifications to let the generated sentences could be perceived as written by a specific author. First, we introduce a convolutional neural network (CNN) as discriminator and two Long Short-Term Memory (LSTM) as the hierarchical structure of generator. In each step of generation process, the discriminator will leak the features of personalized information to generator as a guide signal. The generator accepts this signal to adjust the output when generating the next word. In order to allow GAN to accept personalized information as part of the input, we changed the structure of CNN and the in/output of the generator.
\begin{figure}[h]
	\centering  
	\includegraphics[width=0.7\linewidth]{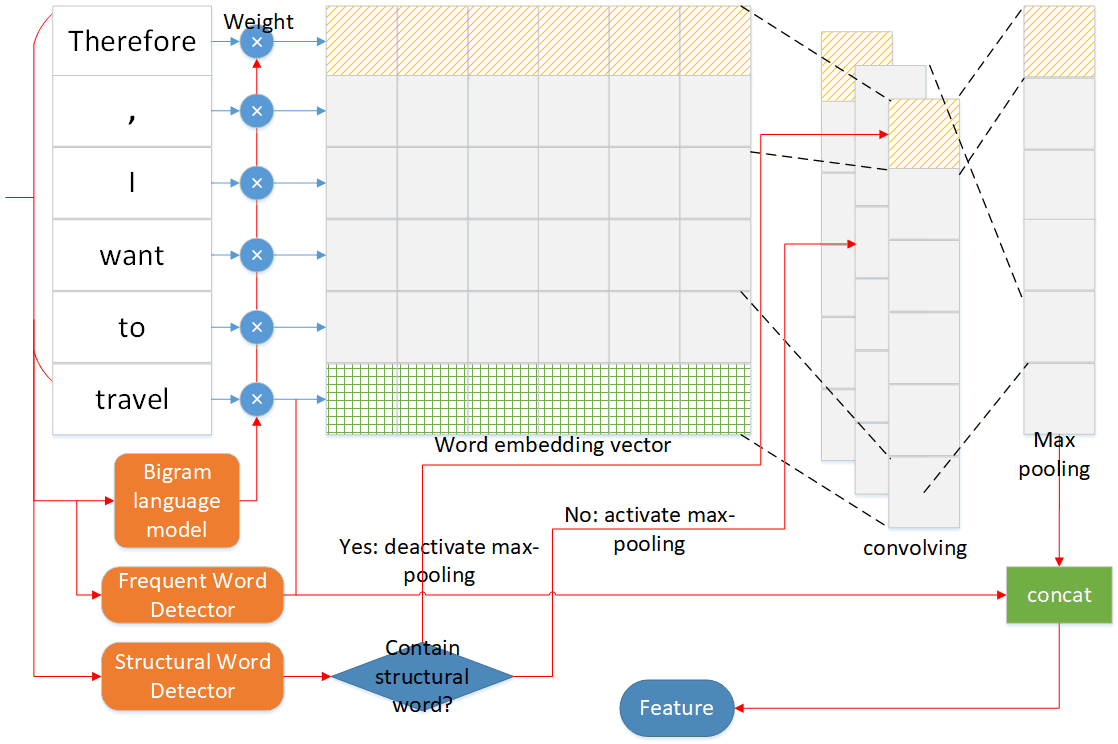}  
	\caption{The flowchart of the modified discriminator CNN.}  
	\label{03}   
\end{figure}
\begin{figure}[h]
	\centering 
	\includegraphics[width=0.7\linewidth]{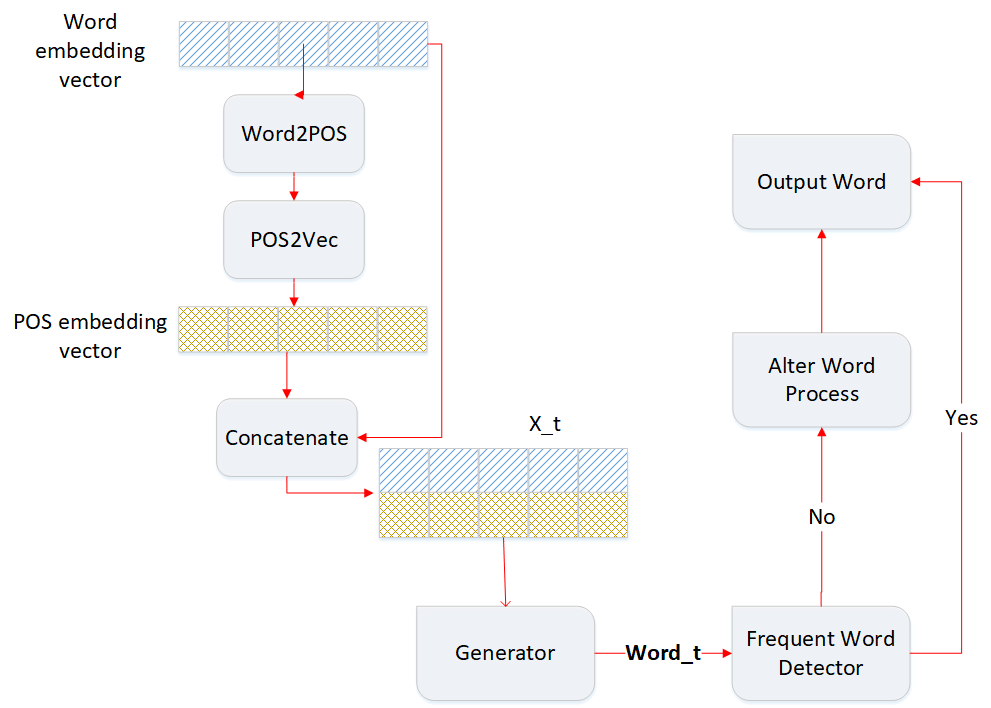} 
	\caption{The flowchart of the modified sentence generator.}  
	\label{04}   
\end{figure}

\subsubsection{DISCRIMINATOR CNN}:\\
\\
The flowchart of the modified discriminator is illustrated in Fig. \ref{03}. To enable the discriminator has the ability to judge whether the generated sentences has the frequently used syntactic structure of the target author, the bigram language model for the target author is applied here. The values of the feature vector is weighted based on the occurrence frequency of the POS tags for input sentences. For example, the POS sequence of the phrase “I want” is “NN VP” and the occurrence frequency of that POS combination in the bigram language model is 0.03, then 0.03 is served as the weight to the word embedding vectors of those two words.\\
\\
Second, for the max-pooling step of the discriminator, the max values of the filters are usually chosen. However, in our goal, the structure words, which represents the syntactic structure information of the target author, should be treated with higher priority. Therefore, when a sentence is input into CNN, our model will first determine whether the sentence contains a structure word. If the sentence contains a structure word, the CNN will select the value corresponding to the window containing the convolution result of the structure word in the max-pooling layer. As a result, the structure word will be stored as the most important value in the max-pooling layer. Otherwise, the CNN model still uses the maximum value in each filter to the max-pooling layer.\\
\\
Other than structure word, the frequently used words of the author is also one important information that represent the personalized writing style. Therefore, the feature leaked from the discriminator should contain this feature to guide the generator. Our model scans for word embedding vector corresponding to these words. If there are frequently used words in the sentence, some of them will be concatenated with output of max-pooling layer. For example, the output dimension of the max-pooling layer is $N\times{M}$, where N is the number of filters and M is the number of window sizes. After concatenating, this matrix becomes $(N+L)\times{M}$, where L is the dimension of the word embedding vector. If there is no word that target author prefer to use in the sentence, then two vectors whose elements are zeroes and are concatenated with output of max-pooling layer.
\subsubsection{GENERATOR}:\\
\\
Based on the LeakGAN, an LSTM module named MANAGER in the Generator accepts guide signal leaked from discriminator and outputs a goal vector. Another LSTM named WORKER module uses the goal vector as the guide signal when generating the next word. As shown in the Fig. \ref{04}, when generating a new word for the sentence, the generator will use the embedding vector of the already generated word sequence as the input conventionally. However, the proposed method add the POS embedding of the word sequence in order to let the syntactic structure of the target author. Therefore, at each generation step, the length of the input vector for WORKER is $2\times{M}$, where M is the dimension of word embedding vector, and $1\times{M}$ is the word embedding vector and the other $1\times{M}$ is the POS vector corresponding to the word sequence\\
\\
Finally, after generating a new sentence, the “alter word process” module is adopted to check the generated sentence and enhance it. If the generated word is a commonly-used word among all authors or the word that the target author preferred, then the generated word will be kept. Otherwise, the word embedding vector for the generated word will be obtained by looking up the pre-trained word embedding, which is generated during the personalized information extraction phase. The closest word that is frequently used by the author will be substituted based on the cosine similarity of the word embedding vector. Then, the sentence generation is done and the generated sentence will be decided whether or not a good generation by the discriminator.
\section{Experimental Analysis}
In order to evaluate the performance of the proposed method, the LeakGAN algorithm is served as the baseline system for comparison. We used the news corpus crawled from the world view section of Washington Post as the experimental corpus. Our model was compared with two kinds of LeakGAN. One is trained with texts of all authors that are qualified NER filter, and the other is trained with all texts of one author. We evaluate the similarity between the generated sentences and the original text of the target author. Another evaluation is focused on the topic relation of the generated sentences and the user-defined topics.
\subsection{Data and Model Preparation}
Because each news reporter has his or her own different writing habits when writing articles, and most of the news events reported by the authors in the same period of time have strong correlation, we use news corpus as the training corpus of our model. We crawl the news covered nearly one year of ten reporters from the “worldviews” section of the Washington Post, which has a total of 1,013 articles. Because the news text needs to quote what others say in order to reflect objectivity, which does not represent the author’s personalized writing habits, we deleted the long quotes. According to our statistics, the distributions of sentence length of all authors fit to the normal distribution, and its mean is 23.01 standard deviation is 11.93, so the maximum length of the sentence we reserved is 46. For better generating sentence, we tokenized the text set, including deleting punctuation marks such as the period and question mark. The total number of words in all articles is 33,961.\\
\\
We set two different size of CNN kernels: 2, 3, and the number of filters for each size is 125. The dimension of the word embedding vector that used as input of CNN is 75. We use dropout as proposed in \cite{fang2015captions}, which maintains at 0.75, and L2 regularization to avoid overfitting. As part of the input to the WORKER module of the generator, the dimension of the word embedding vector in generator is 32, which is the same dimension as the POS embedding vector produced during the personalized information extraction stage. The StanfordNERTagger \cite{finkel2005incorporating} is adopted to extract the named entities that appear in each article. The StanfordPOSTagger \cite{toutanova2000enriching} is also applied to perform the POS tagging of articles to get the occurrence distribution of the bigram structural words. To get word embedding and POS embedding, we use genism \cite{rehurek2010software} to implement the Word2vec algorithm. The Texygen tool \cite{zhu2018texygen} is a benchmark platform that integrates several GAN models for text generation, such as MaliGAN \cite{che2017maximum}, RankGAN \cite{lin2017adversarial}, LeakGAN \cite{guo2018long}, etc. Because Texygen is an open source software, which allows us to change some source codes to perform experiment efficiently, we use the LeakGAN module in Texygen as a baseline system.
\begin{table}
\caption{Personalized sentence similarity for comparing methods.
(Note that lower SimHash score indicates better similarity, and
higher Bleu-3 score is better.)}
\label{tab:table1}
\begin{center}
\begin{tabular}{lrr}
\hline
\bf & \bf SIMHASH & \bf BLEU-3\\
\hline
LEAKGAN & 21.73 & 0.14\\
LEAKGAN-ER & 22.06 & 0.09\\
PERSONGAN & 20.16 & 0.20\\
\hline 
\end{tabular}
\end{center}
\end{table}
\subsection{Personalized Sentence Similarity Evaluation}
\begin{figure}[h]
\begin{minipage}[t]{0.4\linewidth} 
\centering     
\includegraphics[width=1.2\textwidth]{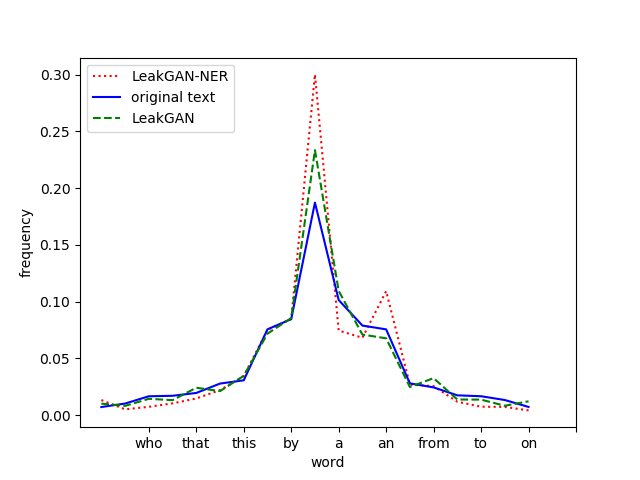}
\caption{Sorted occurrence frequency of structure words for comparing methods and original texts from the target author.}
\label{05}
\end{minipage} 
\hfill
\begin{minipage}[t]{0.4\linewidth}
\centering
\includegraphics[width=1.2\textwidth]{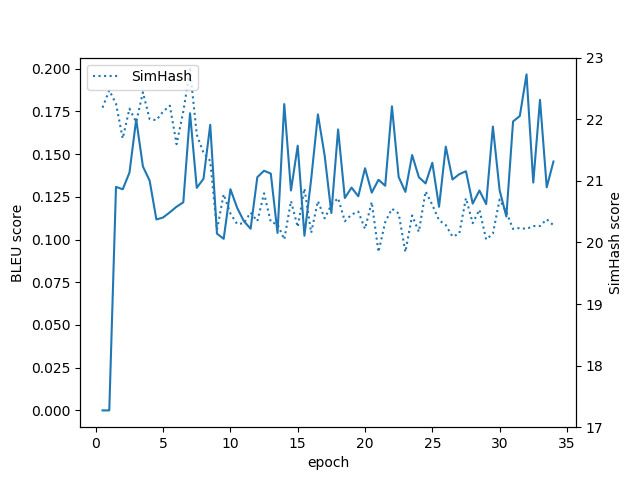}
\caption{The Evaluation of the proposed PersonGAN method for each epoch.}
\label{06}
\end{minipage}
\end{figure}

In the proposed method, the main idea is to let the user decide what topic they want for sentence generation. Therefore, we first compare the difference between baseline system without NER information. The first system is Leak-GAN, which is trained using texts of the target author. The second system is LeakGAN-NER, which is trained using the text set of NER filtered sentences.\\
\\
Here we adopt two popular objective measurements to evaluate the sentence similarity. The first one is
SimHash \cite{charikar2002similarity}. SimHash is one widely used effective way to remove duplicate text. The feature vector of each word is obtained from a given sentence first, and weights are set for each feature vector, which indicates the importance of that vector. Then, the hash value of each feature vector is calculated by a hash function. In this way, the sentence becomes a series of values. The hash value and the weight are positively multiplied when 1 is encountered, and the hash value is multiplied by the negative value when the 0 is encountered. The weighted results of the above respective feature vectors are accumulated to become only one sequence. For the cumulative result of the n-bit signature, if it is greater than 0, it is set to 1, otherwise it is set to 0, so that the simhash value of the sentence is obtained. Finally, we can determine the sentence similarity according to the Hamming distance of SimHash values of different sentences, which is the number of different digits of SimHash value.\\
\\
The second measurement is the Bleu score \cite{papineni2002bleu}. Bleu score measures the fluency and translation quality of generated text by measuring the similarity between the generated text and the reference text. We calculated the unigram, bigram, and trigram language models and give each language model the same weight. The target author’s sentence which contains the NER-related words are served as the reference sentences set to test the average of Bleu and SimHash score. The results are shown in Table \ref{tab:table1}. The Bleu score of LeakGAN-NER is slightly worse than that of Leak-GAN, the possible reason is that when calculating the Bleu score, the reference sentences are collected from the texts of the target author. Since a large portion of the training corpus used by LeakGAN-NER are collected from other authors, the generated sentences would not be similar to the target author; especially the usage of the structure words. We further analysed the occurrence frequencies of the structure words using comparing methods and the original texts from the target author. As shown in Fig. \ref{05}, the occurrence frequency of structural words generated by LeakGAN is more similar to that of the original text than the LeakGAN-NER, which suggests that the syntactic structure of the generated sentences by LeakGAN is more similar to the original text.\\
\\
Based on the previous results, it seems that if the user-defined topic is not collected from the target author, the generated sentence will not similar to the original texts of the target author. Here, we further compare our proposed method with the LeakGAN baseline system, to see under the condition of the user-defined topic, how the performance of the proposed method in terms of sentence similarity. The proposed method is based on the LeakGAN-NER, however, by applying the guiding information such as weighted features from bigram structure word for generator, as discussed in the previous Section, are named PersonGAN. Fig. \ref{06} shows the SimHash score and Bleu score of each training epoch for PersonGAN. The final value of SimHash of PersonGAN is 20:16 and Bleu is 0:20, which are slightly better than the LeakGAN system. It can be seen that the Bleu score has increased significantly after the beginning of the completely randomly generated sentence. And after a few epoch, Bleu score of the text generated by PersonGAN is comparable to that of the LeakGAN and even slighly better as shown in Table 1, which indicates that the proposed method is helpful for guide the GANs to generate sentences similar to the original texts, even though the training texts are written by other authors.
\subsection{Named Entity Recognition Evaluation}
To evaluate topic similarity of the generated sentences by the comparing methods to the user-defined topics, the named entities of those topics are used to find similar words based on the cosine similarity obtained by Word2Vec algorithm. Note that only content words are considered here to evaluate the performance. Then, the occurrence ratio of each sentences is calculated to see if the similar words generated in the sentence. As a result, the ratio is 0.093 in LeakGAN and 0.113 in PersonGAN. These results indicate that the sentence generated by PersonGAN is more topic-related than LeakGAN because the training set of PersonGAN is collected from the topic-related texts.
\section{Conclusion and Future Work}
In this paper, we proposed a personalized sentence generation method for user-defined topic based on GAN framework. By modifying the word embedding feature vector using the guiding information of bigram structure words for generating words, it indeed help generating sentences with similar word usage of the target speaker, even when the training text set is collected from other authors. The named entity recognition is also helpful for generating sentences consist of related words to the user-defined topics. We will further investigate the possibility of generating sentences with higher level sentence hierarchy in order to let the generated sentences could construct a paragraph.
\bibliographystyle{plain}
\bibliography{paper}
\end{document}